\documentclass{article}

\PassOptionsToPackage{sort, compress, numbers}{natbib}

\usepackage[final]{nips_2018}

\usepackage[utf8]{inputenc} %
\usepackage[T1]{fontenc}    %
\usepackage{hyperref}       %
\usepackage{url}            %
\usepackage{booktabs}       %
\usepackage{amsfonts}       %
\usepackage{nicefrac}       %
\usepackage{microtype}      %

\usepackage{amsmath}
\usepackage{amssymb}

\usepackage{graphicx}
\usepackage{subcaption}

\newcommand{\tr}{\ensuremath{ \mathrm{tr}\,}}

\usepackage{color}

\newcommand{\be}{\begin{equation}}
\newcommand{\ee}{\end{equation}}
\newcommand{\bi}{\begin{itemize}}
\newcommand{\ei}{\end{itemize}}

\newcommand{\CC}{\mathcal{C}}
\newcommand{\CA}{\mathcal{A}}
\newcommand{\CB}{\mathcal{B}}
\newcommand{\CN}{\mathcal{N}}
\newcommand{\avg}[1]{\langle #1 \rangle}

\title{SGD Implicitly Regularizes Generalization Error}

\author{
  Daniel A. Roberts\thanks{Work done while at Facebook AI Research in 2018.} \\
  MIT and Salesforce \\
  \texttt{drob@mit.edu} 
  }

\begin{document}

\maketitle

\begin{abstract}
We derive a simple and model-independent formula for the change in the generalization gap due to a gradient descent update. We then compare the change in the test error for stochastic gradient descent to the change in test error from an equivalent number of gradient descent updates and show explicitly that stochastic gradient descent acts to regularize generalization error by decorrelating nearby updates. These calculations depends on the details of the model only through the mean and covariance of the gradient distribution, which may be readily measured for particular models of interest. We discuss further improvements to these calculations and comment on possible implications for stochastic optimization.\\~\\
\textbf{Note added}: this paper appeared in the ``Workshop on Integration of Deep Learning Theories'' at NeurIPS in 2018 \cite{roberts2018sgd}. Given the current interest in this topic (see e.g. \cite{smith2021on}), we would like to make the workshop paper more widely available. Other than this note and an updated affiliation, the paper is otherwise unchanged.

\end{abstract}

\section{Introduction}

Deep learning models are almost all trained using gradient-based learning algorithms \cite{bottou2016optimization}. Remarkably, these algorithms continue to work well even in cases where the models have far more free parameters than the number of training examples that are used to learn their value. While it's not shocking that such models with many degrees of freedom can learn on a fixed training set \cite{zhang2016understanding}, it's surprising that they also generalize exceedingly well on novel examples. 

Furthermore, stochastic gradient descent (SGD) \cite{robbins1951stochastic}, in which only a small subset or mini-batch of examples is used to estimate the gradient, is not only known to lead to configurations that generalize well \cite{bottou2005line,bottou2008tradeoffs}, but in many of these large-scale modeling scenarios will far outperform full-batch gradient descent (GD) despite estimating the gradient with fewer examples \cite{keskar2016large,SmithLeBayes}. 

A central question is whether this effectiveness is due to some very general property of deep learning or instead whether it's due to the dynamics of the learning algorithm used to train them. It seems that these algorithms must employ some kind of \emph{implicit regularization} \cite{neyshabur2017geometry}, since such generalization performance does not really degrade as model size increases.

The prevailing explanation given for this empirical observation has to do with descriptions of the loss landscape \cite{keskar2016large} (also see e.g. \cite{SmithLeBayes,chaudhari2016entropy,jastrzkebski2017three,hu2017diffusion,kleinberg2018alternative,zhu2018regularization}), in particular that SGD prefers ``flat'' to ``sharp'' minima in which the loss doesn't change too much in the neighborhood of the minima. An argument originating from \cite{hochreiter1997flat} ascribes good generalization properties to such flat minima, reasoning that the effect of changing inputs could be reinterpreted as shifting the location of the minima. (Although due to the activations having rescaling symmetries \cite{neyshabur2015path}, this should be understood to apply only after such reparameterizations have been taken into account \cite{dinh2017sharp}.)

A common mechanism offered for this behavior is that the stochasticity is equivalent to injecting noise into the optimization process, allowing for barriers in the landscape to be climbed and sharp minima to be avoided \cite{keskar2016large}. However, isotropic noise to full-batch GD does not help for generalization (though it might for SGD \cite{neelakantan2015adding}?) suggesting that it's anisotropy that is important \cite{ZhangEnergyEntropy}, and \cite{zhu2018regularization} argues that anisotropic noise is necessary for escaping from sharp minima. Either way, most of these analyses rely on taking the continuum limit \cite{hu2017diffusion,mandt2017stochastic,chaudhari2017stochastic,jastrzkebski2017three,lampinen2018analytic,ZhangEnergyEntropy} in a way that is not necessarily justified,\footnote{The continuum limit implicitly assumes that the mini-batch size vanishes faster than the learning rate, despite the batch size being bounded from below by 1. For further discussion, see section 2.3.3 of \cite{yaida2018fluctuation}.} and moreover assume that the covariance of the gradient does not depend strongly on the value of the model parameters. As a result, this leads many authors to conclude that some kind diffusion process in parameter space must play a fundamental role in finding minima that generalize well \cite{chaudhari2017stochastic,jastrzkebski2017three}.

Here, we attempt to shed light on this implicit regularization puzzle by shifting focus away from the loss landscape and instead considering the discrete update dynamics themselves. Our analysis is very general and does not make any assumptions about the model or dataset.\footnote{Cf. \cite{lampinen2018analytic}, which focuses on deep linear networks in the student-teacher scenario in the continuum limit.}  Instead, we derive formulas for the change in generalization error after an update and for the difference between GD and SGD gradients in terms of readily measurable quantities that characterize the gradient distribution.

Our key result is that through accumulated updates GD develops a ``bias'' toward overfitting, which SGD mitigates by decorrelating nearby updates.\footnote{See \cite{yin2017gradient} in which the role of ``gradient diversity'' for generalization is also highlighted, although it is not connected to the different dynamics between GD and SGD as we do here.} 
After averaging over the datasets, we show that the SGD gradient differs from the GD gradient by the derivative of the change in the generalization error to leading order in the learning rate 
\be
\avg{\delta g }\propto  \partial \avg{\delta \epsilon},\label{eq:SGD-main-result}
\ee
where $\avg{\delta \epsilon}$ is the average change in the generalization error, and $\avg{\delta g}$ is the averaged difference between SGD and GD gradients. Furthermore, we show that $\avg{\delta \epsilon}$ is proportional to the trace of the gradient covariance matrix $\Sigma$
\be
\avg{\delta \epsilon} \propto \tr \Sigma,
\ee
which means we can interpret \eqref{eq:SGD-main-result} as making ``explicit'' the implicit regularization of SGD over GD. This also means that these results would be missed in any analysis that considers a homogeneous covariance independent of the model parameters as $\avg{\delta g }$ would vanish identically. Although we omit experiment results for brevity, it's easy to check that $\partial\, \tr \Sigma$ tends to become very large over the course of training, and so we think that this ``bias'' is more central to the generalization puzzle than diffusion. The upshot is that SGD noise not only leads to diffusion, but also to drift. 

In this extended abstract, we will report only on our initial theoretical results, holding experimental verification for the completed work.

\section{Gradient Descent as Stochastic Optimization}\label{sec:gd-as-stochastic}
The update equation for the parameter vector in both GD and SGD can be written as 
\be
\theta_{t+1}^i = \theta_t^i - \eta \, g_i^{(\CB)}(\theta_t), \qquad g_i^{(\CB)}(\theta) \equiv \frac{1}{|\CB|} \sum_{b=1}^{|\CB|} g_i(\theta, x^{(b)}),\label{eq:update}
\ee
where $\theta^i_t$ is the parameter vector after $t$ updates with the index $i$ ranging over all the parameters, $g_i(\theta, x)$ is the gradient with respect to the parameters evaluated at $\theta$ and example $x$, and $\eta$ is the learning rate. We will use the shorthand $g_i^{(\CB)}(\theta)$ to indicate the sample average of a gradient on a batch of examples $\CB$. If $\CB$ is the entire training set, then \eqref{eq:update} is the full-batch gradient descent update, but if $\CB$ is a subset of the training set, then \eqref{eq:update} is an SGD update.

With a shift in perspective, we can treat GD and SGD on the same footing. If we consider averages over all possible realizations of the training set, then both GD and SGD are stochastic, differing just in which examples are used as estimators for the gradient. Ultimately, we are interested in learning a distribution $p(x)$ from which our training and test examples are sampled, and we are not at all attached to the examples in our particular training set.

Furthermore, we see that the only way the input $x$ influences the dynamics is through the batch average of the gradient $g_i^{(\CB)}(\theta)$. Thus, regardless of how complicated the input distribution is, we never have to characterize it directly. Instead, we can just focus on the induced gradient distribution.

In general, the gradient distribution $p(g|\theta,x)$ will also be a complicated distribution, with the higher moments playing an important role. (The number of parameters required to describe such a distribution will be exponential in the number of parameters required to represent the random variable $g$.) However, considering \eqref{eq:update}, we see that $g_i^{(\CB)}(\theta)$ is a sum over many independent random variables $g_i(\theta, x^{(b)})$. This means that for large enough batch size the central limit theorem guarantees that the distribution of the gradient estimator $g_i^{(\CB)}(\theta)$ will be Gaussian. Using the notation $\avg{\cdot} \equiv \int dx \, p(x) \,\cdot$ to indicate an average over the data generating distribution, the gradient mean $G_i(\theta)$ and gradient covariance matrix $\Sigma_{ij}(\theta)$ are given by
\be
G_i(\theta) \equiv \avg{g_i(\theta,x)}, \qquad  \delta^{ab}\Sigma_{ij}(\theta)  \equiv \avg{ g_i (\theta, x^{(a)}) \, g_j (\theta, x^{(b)})} - G_i(\theta) \, G_j(\theta),
\ee
where $ \delta^{ab}$ indicates that gradients derived from different independent examples have no covariance.
Then, the joint distribution of two gradient estimators $g_i^{(\CA)}(\theta)$ and $g_i^{(\CB)}(\theta)$  is Gaussian 
\be
    p(g_i^{(\CA)}, g_j^{(\CB)}|\theta) \sim \CN\bigg(G_i(\theta), G_j(\theta) ; \frac{|\CA \cap \CB|}{|\CA| \, |\CB|}\Sigma_{ij}(\theta) \bigg), \label{eq:distribution}
\ee
where the covariance is the covariance on a single gradient scaled by the size of the intersection of the two sets, divided by the size of both sets. Crucially, the noise $\Sigma_{ij}$ is anisotropic and depends on $\theta$.

Now, let's use the distribution \eqref{eq:distribution} to compute the change in generalization error incurred by a gradient descent update. A GD update induces a change in the parameters $\delta \theta^{i\,(\CB)} \equiv - \eta \, g_i^{(\CB)}$.
Since we don't actually get to observe the loss directly, we also need a batch of samples from $p(x)$ to make an estimator. Let $\ell(\theta, x)$ be the loss on a particular example and define $\ell^{(\CA)}(\theta) \equiv \frac{1}{|\CA|}\sum_{a=1}^{|\CA|} \ell(\theta, x^{(a)})$
as the loss estimator on a batch $\CA$. Let us also define the change in the loss estimator after an update $\delta \theta^{i\,(\CB)}$ as
$\delta \ell^{(\CA, \CB)}(\theta) \equiv \ell^{(\CA)}(\theta + \delta \theta^{(\CB)}) -\ell^{(\CA)}(\theta)$.
After an update \eqref{eq:update}, the loss estimator changes by an amount
\be
\delta \ell^{(\CA, \CB)} = -\eta \, g^{(\CA)} \cdot g^{(\CB)} + O(\eta^2),
\ee
where we have expanded in the learning rate $\eta$ and suppressed the dependence on $\theta$. 

Taking an expectation over $p(x)$, we find that on average our loss estimator decreases to first order in $\eta$, with the amount determined by the second moment of the batch-gradient distribution \eqref{eq:distribution}
\be
\avg{\delta \ell^{(\CA, \CB)} } = -\eta \bigg[||G||^2 +  \frac{|\CA \cap \CB|}{|\CA| \, |\CB|}\tr  \Sigma \bigg].
\ee
The terms in the square brackets are manifestly positive; the loss manifestly decreases at order $\eta$. The first term, the norm squared of the mean of the gradient, is the expected decrease of an objective function from a gradient descent update. The second term, proportional to the trace of the gradient covariance matrix, is an enhancement that comes from the fact that the loss estimator and gradient estimator might have a positive correlation. 

Now, if we let $\CA$ be a test set and $\CB$ be a training set of $N$ non-overlapping examples, we find
\be
\avg{\delta\ell}_{test} \equiv \avg{\delta \ell^{(\CA, \CB)} } =  - \eta \, ||G||^2, \qquad \avg{\delta\ell}_{train} \equiv \avg{\delta \ell^{(\CB, \CB)} } = - \eta \, ||G||^2 - \eta \frac{\tr  \Sigma}{N}.
\ee
Defining the change in the generalization gap to be the difference between the change in the test error compared to the training error, $\delta \epsilon \equiv \delta\ell_{test} - \delta\ell_{train}$,
this quantity and its average are given by
\be
\delta \epsilon = \eta \, g^{(\CB)} \cdot (g^{(\CB)} - g^{(\CA)}) + O(\eta^2), \qquad \avg{\delta \epsilon}  = \eta \frac{\tr  \Sigma}{N} + O(\eta^2).\label{eq:gd-ge}
\ee
The formula for the average change in the generalization gap $\avg{\delta \epsilon}$ is rather sensible and even intuitive. To leading order in $\eta$, the difference between the test and training error is proportional to the learning rate, inversely proportional to the size of the training set, and related to the fluctuations of the gradient across different data realizations through the trace of the covariance matrix. First, if the training set was infinitely large $(N \to \infty)$, the training set would capture the full variance of the data generating distribution and there would be no generalization gap. %
Second, if there were no variance in the gradient, then the training set and test set would lead to the same updates and not overfit.

\section{Implicit Regularization}
\label{sec:sgd}

The generalization gap from gradient descent arose because the training set is a biased estimator of the average loss, and we showed that the bias can be computed explicitly \eqref{eq:gd-ge}. In this section, we will compare stochastic gradient descent directly with gradient descent. In order to see a meaningful difference, we need to consider more than one update.

Consider two subsequent SGD updates with minibatches $\CB$ and $\CC$ each containing $N/2$ examples
\be
\theta_1^{i\, (\CB)} \equiv \theta_0^i - \eta \, g_i^{(\CB)}(\theta_0), \qquad \theta_2^{i\, (\CB, \CC)} \equiv \theta_1^{i\, (\CB)} - \eta \, g_i^{(\CC)}(\theta_1).
\ee
Plugging the first equation into the second and Taylor expanding, we find that the parameters after two updates $\delta \theta_2^{i\, (\CB, \CC)} \equiv \theta_2^{i\, (\CB, \CC)} - \theta_0^i$ can be expressed as
\be
\delta\theta_2^{i\, (\CB, \CC)} = - \eta \Big[g_i^{(\CB)} + g_i^{(\CC)} \Big] + \eta^2 g_j^{(\CB)} h_{ij}^{(\CC)} + O(\eta^3), \label{eq:two-updates}
\ee
where the gradients and Hessians are all evaluated at the initial point $\theta_0$, $h_{ij}^{(\CC)}$ is the Hessian estimator $\frac{1}{|\CC|}\sum_{c=1}^{|\CC|} \partial_i \partial_j \ell(\theta,x^{(c)})$, and we use the convention that repeated indices are summed over.

From this alone, we can see that the parameter updates for GD and SGD will be different. Let's express this in terms of the difference between the overall GD gradient and the overall SGD gradient after performing two updates so $\delta\theta_2^{i\, (\CB\cup\CC, \CB\cup\CC)}- \delta\theta_2^{i\, (\CB, \CC)} \equiv -\eta \, \delta g_i$. Then, making use of the fact that we can rewrite $h_{ij} g_j = \partial_i g^2 /2$, we see that this difference in gradients is given by
\be
\delta g_i = -\frac{\eta}{8} \Big[ \partial_i \, ||g^{(\CB)} - g^{(\CC)}||^2 + 4 \big(g_j^{(\CC)} h_{ij}^{(\CB)} -  g_j^{(\CB)} h_{ij}^{(\CC)}\big)  \Big].
\ee
Averaging over different data realizations and substituting in \eqref{eq:gd-ge}, we see that
\be
\avg{\delta g_i } = -\frac{1}{2} \partial_i \avg{\delta \epsilon}.\label{eq:sgd-ge}
\ee

This is our main result. To leading order in $\eta$, the difference between SGD and GD can be interpreted in terms of a difference in the gradient vector equal to the derivative of change in the average generalization error.
By considering a modified loss
$\tilde{\ell}(\theta, x) \equiv \ell(\theta, x) + \frac{1}{4} \avg{\delta \epsilon}$, up to $O(\eta^3)$ this means that two full-batch GD updates of size $N$ each can emulate two non-overlapping half-batch SGD updates of size $N/2$ each, making the \emph{implicit regularization} of \cite{neyshabur2017geometry} explicit.

\section{Various Comments}\label{sec:comments}
Many generalizations were held out for brevity. For instance, the calculation in \S\ref{sec:sgd} is simple to extend to an epoch of any number of updates, and the conclusion remains the same (though the coefficient is different). We can also modify the calculations for second-order methods or to include momentum. 

Our results \eqref{eq:gd-ge} and \eqref{eq:sgd-ge} are model-independent formula that relate performance metrics of interest to simple quantities that can be measured empirically for any particular model and dataset of interest. However, for some models it's possible to evaluate these quantities directly, e.g. for a linear model with quadratic loss and Gaussian input data. 
In that case, we can show explicitly that in this context SGD's implicit regularization can behave like $L_2$ regularization of the parameter vector.

Since our results are essentially expressed in terms of a dual expansion in $\eta$ and $1/N$ (or $1/$batch-size), there are many corrections to discuss, such as next to leading order in $\eta$, or whether the quantities $\delta \epsilon$ and $\delta g_i$ self-average. %
These calculations are easily performed, but were also held for brevity. 

One important correction concerns the fact that the gradient distributions evaluated on the test set vs. the training set will begin to diverge quickly as a model trains and overfits.\footnote{This is easy to measure and could serve as a new statistic for overfitting that is worth investigating further.} Considering \eqref{eq:gd-ge}, if these distributions are different, then the change in generalization error would become
$ \eta \, \tr \Sigma_{train}/N + \eta\, G_{train} \cdot( G_{test} - G_{train})$,
and thus has an additional piece that depends on the difference between test and train mean gradients. %
Since the second term is still present, this doesn't falsify our result \eqref{eq:sgd-ge}, but shows that this divergence of gradients is also an important factor in overfitting.

Finally, we note that these results can lead to improved stochastic optimization algorithms. For instance, to leading order one can go to larger batch sizes (perhaps even full-batch) by modifying the loss to make the SGD regularization explicit. Or one could use the inverse of $\Sigma_{ij}$ as a preconditioner to slow down learning in directions that the gradient is strongly fluctuating.\footnote{In general $\Sigma_{ij}$ is very anisotropic, and it's exceptionally computationally efficient to compute its leading eigenvalues and eigenvactors. Together, these facts make such an algorithm very practical.} This latter method appears to work rather well, and we hope to report on it in an extended version of the present work.

\section*{Acknowledgments}
We are grateful to L\'eon Bottou, Boris Hanin, Max Kleiman-Weiner, Yann LeCun, Tengyu Ma, DJ Strouse, and Sho Yaida for comments and discussions.
This extended abstract was brought to you by the letter $\Sigma$ after averaging over many different realizations.

\bibliographystyle{unsrtnat}
\bibliography{generalization}{}

\end{document}